\documentclass[acmsmall, nonacm, screen]{acmart}

\title[Application of NotebookLM for Lung Cancer Staging]{Application of NotebookLM, a Large Language Model with Retrieval-Augmented Generation, for Lung Cancer Staging}

\makeatletter
\g@addto@macro\@authornotes{\footnotetext{(1) Department of Radiology, University of Yamanashi, Yamanashi, Japan.}}
\g@addto@macro\@authornotes{\footnotetext{(2) Department of Radiation Oncology, Tohoku University Graduate School of Medicine, Sendai, Japan.}}
\makeatother

\author{Ryota Tozuka$^\text{(1)(2)}$}
\author{Hisashi Johno$^\text{(1)}$}
\authornote{
  Corresponding author: Hisashi Johno, M.D., Ph.D.
  Department of Radiology, University of Yamanashi
  1110 Shimokato, Chuo, Yamanashi, 409-3898, Japan.
  Tel: +81-55-273-1111, Fax: +81-55-273-6744.
  E-mail: johnoh@yamanashi.ac.jp.
}
\author{Akitomo Amakawa$^\text{(1)}$}
\author{Junichi Sato$^\text{(1)}$}
\author{Mizuki Muto$^\text{(1)}$}
\author{Shoichiro Seki$^\text{(1)}$}
\author{Atsushi Komaba$^\text{(1)}$}
\author{Hiroshi Onishi$^\text{(1)}$}
\authorsaddresses{}
\thanks{This study was partially supported by JSPS KAKENHI Grant Number JP21K15762.}

\usepackage{enumitem}
\usepackage[capitalize]{cleveref}
\usepackage[range-units=single]{siunitx}
\usepackage{tikz}
\usetikzlibrary{calc, positioning}
\usepackage{titlesec}
\usepackage{sepfootnotes}

\titleformat{\section}{\Large\bfseries}{\thesection}{1em}{}

\begin{document}
  \maketitle
  \renewcommand{\shortauthors}{Tozuka et al.}

  \section*{Abstract}
  \subsection*{Purpose}
  In radiology, large language models (LLMs), including ChatGPT, have recently gained attention, and their utility is being rapidly evaluated. However, concerns have emerged regarding their reliability in clinical applications due to limitations such as hallucinations and insufficient referencing. To address these issues, we focus on the latest technology, retrieval-augmented generation (RAG), which enables LLMs to reference reliable external knowledge (REK). Specifically, this study examines the utility and reliability of a recently released RAG-equipped LLM (RAG-LLM), NotebookLM, for staging lung cancer.

  \subsection*{Materials and methods}
  We summarized the current lung cancer staging guideline in Japan and provided this as REK to NotebookLM. We then tasked NotebookLM with staging 100 fictional lung cancer cases based on CT findings and evaluated its accuracy. For comparison, we performed the same task using a gold-standard LLM, GPT-4 Omni (GPT-4o), both with and without the REK.

  \subsection*{Results}
  NotebookLM achieved \qty{86}{\percent} diagnostic accuracy in the lung cancer staging experiment, outperforming GPT-4o, which recorded \qty{39}{\percent} accuracy with the REK and \qty{25}{\percent} without it. Moreover, NotebookLM demonstrated \qty{95}{\percent} accuracy in searching reference locations within the REK.

  \subsection*{Conclusion}
  NotebookLM successfully performed lung cancer staging by utilizing the REK, demonstrating superior performance compared to GPT-4o. Additionally, it provided highly accurate reference locations within the REK, allowing radiologists to efficiently evaluate the reliability of NotebookLM's responses and detect possible hallucinations. Overall, this study highlights the potential of NotebookLM, a RAG-LLM, in image diagnosis.

  \section*{Keywords}
  Large language model (LLM),
  Retrieval-augmented generation (RAG),
  Reliable external knowledge (REK),
  NotebookLM,
  GPT-4 Omni (GPT-4o),
  Lung cancer staging

  \section{Introduction}
  Large language models (LLMs) are artificial intelligence (AI) models designed to understand natural language and generate human-like responses, achieving considerable success in a variety of domains \cite{Radford}. In radiology, the potential of LLMs has recently been explored in areas such as image diagnostic support, radiology education, and medical physics \cite{Keshavarz,Suzuki,Kadoya}. While these studies emphasize the promising role of LLMs in the field, it has been observed that their responses sometimes deviate from user expectations or fail to align with established clinical consensus. LLMs often process information indiscriminately from diverse references, occasionally generating answers based on unreliable or outdated references. This phenomenon, referred to as LLM hallucinations, can generate seemingly plausible but factually incorrect information, making it difficult to rely on LLMs for clinical diagnoses \cite{Keshavarz,Ji,Huang}.

  Retrieval-augmented generation (RAG) is an emerging technology designed to reduce hallucinations in LLMs by enabling them to reference reliable external knowledge (REK) \cite{Lewis,Shuster}. A key feature of RAG is its ability to link LLM responses to specific sources within REK, allowing users to easily verify the reliability of those responses. Research on RAG-equipped LLMs (RAG-LLMs) in the medical field is still in its early stages \cite{Ge,Zhou,Miao,Kresevic,Mashatian}, and their effectiveness, particularly in image diagnosis, remains largely unexplored.

  NotebookLM (\url{https://notebooklm.google}) is a web-based RAG-LLM, experimentally released by Google Inc. in June 2024. It generates responses by clearly citing sources from user-provided REK, utilizing the advanced language model, Gemini 1.5 Pro \cite{GeminiTeam}. NotebookLM is designed to be user-friendly, requiring no specialized knowledge of AI---users simply need to prepare and provide REK.

  In this study, we focus on the utility of NotebookLM for lung cancer staging. Specifically, we provided NotebookLM with a lung cancer staging guideline as REK and tasked it with diagnosing TNM classifications based on fictional lung cancer CT findings. Its accuracy was then compared to that of GPT-4 Omni (GPT-4o; OpenAI Inc., San Francisco, CA, USA), a widely used LLM.
  \section{Materials and methods}
\begin{figure}
    \includegraphics[height = 20em]{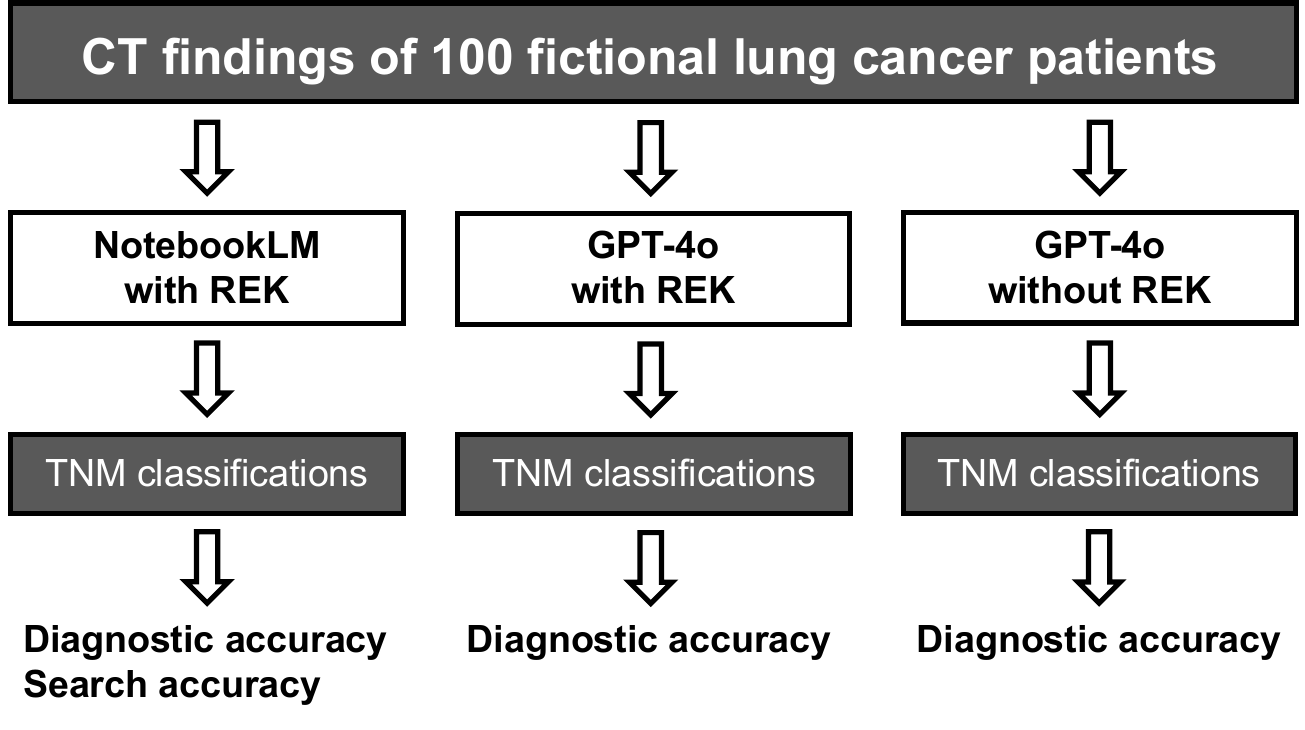}
    \caption{An overview of the experimental process. Radiologists from our team created CT findings for 100 fictional lung cancer patients, and each patient's TNM classification was diagnosed by the different LLM settings (NotebookLM with REK, GPT-4o with REK, and GPT-4o without REK). Our team's radiologists evaluated these diagnoses and calculated their diagnostic accuracies. For NotebookLM, since it searches and explicitly presents reference locations within REK as the basis for its answers, we also assessed the appropriateness of these locations (search accuracy). \textit{REK}=reliable external knowledge.}
    \label{fig1}
  \end{figure}

  An overview of the experimental process is schematically summarized in \cref{fig1}.

  \subsection*{Data preparation}
  \renewcommand{\thefootnote}{\fnsymbol{footnote}}
  Two radiologists from our team generated CT findings for 100 fictional lung cancer patients, along with TNM classifications based on the latest lung cancer staging guideline in Japan, i.e., the 8th edition of the General Rule for Clinical and Pathological Record of Lung Cancer \cite{theJapanLungCancerSociety}. The CT findings and TNM classifications were subsequently reviewed and confirmed by two additional radiologists. The entire process was verified by a senior radiologist. The breakdown of TNM classifications for the 100 fictional lung cancer patients is listed in \cref{table1}. All the fictional CT findings with TNM classifications are available in Online Resource 1\footnote[2]{Online Resources 1 to 3 can be found in the ancillary files uploaded with this paper on arXiv.}, with the first of the 100 cases provided below as an example:
  \begin{quote}
    Case 1 CT findings: {\ttfamily A 4.9 cm tumor with invasion of the great vessels is identified in the left lower lobe. Enlarged lymph nodes are seen in the contralateral hilum. No distant metastasis.}

    Case 1 TNM classification: T4 N3 M0.
  \end{quote}

  \begin{table}
    \caption{The breakdown of TNM classifications for the 100 fictional lung cancer patients.}
    \label{table1}
    \begin{tabular}{l}
    \begin{tabular}{c*{11}{|c}}
      T classification & TX & T0 & Tis & T1mi & T1a & T1b & T1c & T2a & T2b & T3 & T4 \\
      \hline
      Number of patients & 2 & 2 & 4 & 4 & 4 & 9 & 8 & 18 & 10 & 14 & 25 \\
    \end{tabular} \\[1em]
    \begin{tabular}{c*{4}{|c}}
      N classification & N0 & N1 & N2 & N3 \\
      \hline
      Number of patients & 48 & 15 & 19 & 18
    \end{tabular} \\[1em]
    \begin{tabular}{c*{4}{|c}}
      M classification & M0 & M1a & M1b & M1c \\
      \hline
      Number of patients & 60 & 11 & 17 & 12
    \end{tabular}
    \end{tabular}
  \end{table}

  \subsection*{Preparation of REK and prompt texts}
  As REK for NotebookLM and GPT-4o, we summarized the current lung cancer staging guideline in Japan. This REK is available in Online Resource 2\footnote[2]{Online Resources 1 to 3 can be found in the ancillary files uploaded with this paper on arXiv.}. NotebookLM was provided with the REK along with the following prompt ([file name] is the file provided to NotebookLM as REK):
  \begin{quote}
    \ttfamily
    According to {\normalfont [file name]}, identify the TNM classification corresponding to the following CT findings. Note that, for T1 and T2, specify the appropriate subclass (the subclasses for T1 are T1mi, T1a, T1b, T1c; and for T2, T2a and T2b). In addition, the tongue ward refers to a segment of the left upper lobe of the lung. For a primary tumor in the left lung, `ipsilateral' refers to the left side and `contralateral' refers to the right side, whereas for a primary tumor in the right lung, `ipsilateral' refers to the right side and `contralateral' refers to the left side. If a tumor nodule is found in the contralateral lung from the primary lung cancer, it meets the criteria for M1a. If the solid component diameter is less than or equal to 1cm and the overall lesion (ground-glass nodule) diameter is greater than 3 cm, then it meets the criteria for T1a. The right upper lobe, right middle lobe, right lower lobe, left upper lobe, and left lower lobe are distinct from one another.
  \end{quote}
	We used almost the same prompt when providing GPT-4o with the REK. Note that the REK for GPT-4o was supplied directly within the prompt, rather than through RAG. For GPT-4o without REK, the prompt only instructed it to classify the lung cancer stages based on the latest lung cancer staging guideline in Japan.

  \subsection*{Evaluation}
  Diagnostic accuracy for each LLM setting (i.e., NotebookLM with REK, GPT-4o with REK, or GPT-4o without REK) was calculated as the percentage of correctly diagnosed TNM classifications across the 100 lung cancer cases. Note that a TNM classification was considered correct when all the T, N, and M factors were correctly diagnosed. Additionally, diagnostic accuracies of the LLMs for each of the T, N, and M factors were also calculated.

  For NotebookLM with REK, search accuracy was also calculated as the percentage of lung cancer cases, out of 100, in which NotebookLM referenced the appropriate locations within the REK. The correctness of these referenced locations was evaluated by radiologists from our team. In contrast, for GPT-4o with and without REK, the reference points were generally unclear, making it impossible to calculate search accuracy.

  \begin{figure}
    \includegraphics[height = 20em]{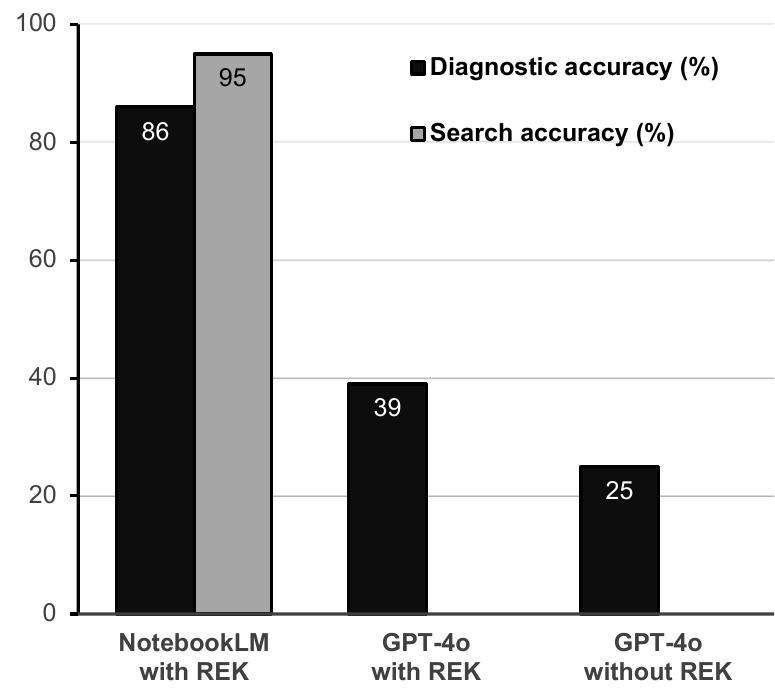}
    \caption{Diagnostic accuracies of TNM classifications for each LLM setting (i.e., NotebookLM with REK, GPT-4o with REK, or GPT-4o without REK) in the experiment using 100 fictional lung cancer cases. A diagnosis of TNM classification was considered correct if all the T, N, and M factors were correctly diagnosed. For NotebookLM with REK, search accuracy was also calculated as the percentage of lung cancer cases in which NotebookLM referenced the appropriate locations within the REK. \textit{REK}=reliable external knowledge.}
    \label{fig2}
  \end{figure}

  \begin{figure}
    \begin{tikzpicture}
      \node[anchor=north west] (input) {Case 2 CT findings:};
      \node[below=0 of input.south west, draw, anchor=north west, text width = .72 \textwidth, xshift=1em] (input text) {\vbox{\ttfamily A 2 cm solid component diameter tumor is observed in the left lower lobe. The tumor invades the main bronchus but not the carina. Enlarged left mediastinal lymph nodes. There is a metastasis in the axial vertebrae.}};
      \node[right=1em of input text, align=left]{Correct answer:\\T2a N2 M1b};
      \path (0, 0 |- input text.south west) + (0, -1ex) coordinate (input bottom);
      \node[below=1ex of input bottom, anchor=north west] (ans) {Output by LLMs:};
      \node[below=0 of ans.south west, anchor=north west, xshift=1em] (nlm) {NotebookLM with REK};
      \node[below=0 of nlm.south west, draw, anchor=north west, text width=.75 \textwidth] (nlm out) {\vbox{
        \ttfamily
        \begin{itemize}[leftmargin=*]
          \item \textbf{T2a:} The tumor size is 2 cm in diameter and invades the main bronchus. According to source \textcircled{\scriptsize 1}, tumors with a solid component diameter greater than 3 cm and less than or equal to 5 cm, or a solid component diameter less than or equal to 3 cm with involvement of the main bronchus meet the criteria for T2. Additionally, source \textcircled{\scriptsize 2} states that tumors meeting the T2 criteria but are 3 cm or less in size are classified as \textbf{T2a}.
          \item \textbf{N2:} The case presents with enlarged left mediastinal lymph nodes. Since the primary tumor is in the left lung, the mediastinal lymph nodes are considered ipsilateral. Source \textcircled{\scriptsize 3} states that metastasis to the ipsilateral mediastinal lymph nodes falls under the \textbf{N2} classification.
          \item \textbf{M1b:} The case notes "metastasis in the axial vertebrae," indicating metastasis outside the lungs. Source \textcircled{\scriptsize 4} states that a single metastasis outside the lungs is classified as \textbf{M1b}.
        \end{itemize}
      }};
      \node[right=1em of nlm out] {Correct};
      \node[below=1ex of nlm out.south west, anchor=north west] (w) {GPT-4o with REK};
      \node[below=0 of w.south west, draw, anchor=north west, text width=.75 \textwidth] (w out) {\vbox{
        \ttfamily
        \begin{itemize}[leftmargin=*]
          \item \textbf{T2b:} Tumor is 2 cm in diameter, invades the main bronchus without carina involvement.
          \item \textbf{N2:} Enlarged left mediastinal lymph nodes.
          \item \textbf{M1b:} Metastasis in the axial vertebrae.
        \end{itemize}
      }};
      \node[right=1em of w out] {Incorrect};
      \node[below=1ex of w out.south west, anchor=north west] (wo) {GPT-4o without REK};
      \node[below=0 of wo.south west, draw, anchor=north west, text width=.75 \textwidth] (wo out) {\vbox{
        \ttfamily
        \begin{itemize}[leftmargin=*]
          \item \textbf{T3:} Tumor invades the main bronchus but not the carina.
          \item \textbf{N2:} Enlarged left mediastinal lymph nodes.
          \item \textbf{M1c:} Metastasis in the axial vertebrae.
        \end{itemize}
      }};
      \node[right=1em of wo out] {Incorrect};
    \end{tikzpicture}
    \caption{A representative result from the lung cancer staging experiment using LLMs. The sources \textcircled{\scriptsize 1} to \textcircled{\scriptsize 4} referenced in the answer by NotebookLM with REK are available in Online Resource 3\protect\footnotemark[2]. \textit{LLM}=large language model, \textit{REK}=reliable external knowledge.}
    \label{repr}
  \end{figure}

  \footnotetext[2]{Online Resources 1 to 3 can be found in the ancillary files uploaded with this paper on arXiv.}

  \begin{figure}
    \includegraphics[height = 20em]{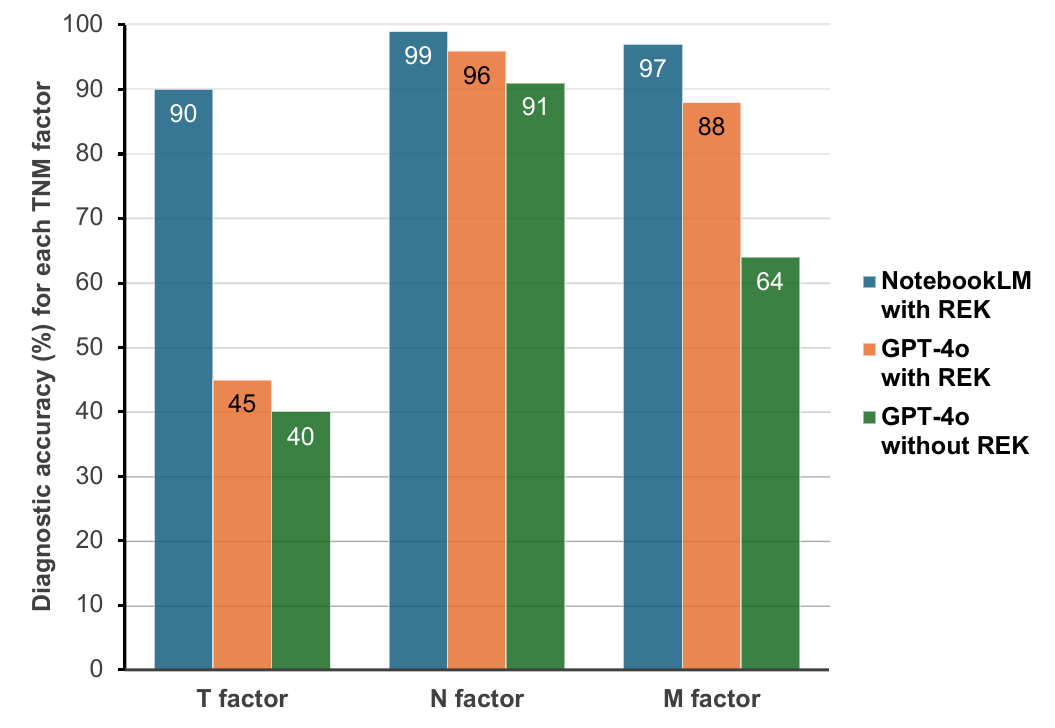}
    \caption{Diagnostic accuracies of the LLMs (NotebookLM with REK, GPT-4o with REK, and GPT-4o without REK) for each of the T, N, and M factors in the experiment with 100 fictional lung cancer patients. \textit{REK}=reliable external knowledge.}
    \label{fig3}
  \end{figure}

  \begin{figure}
    \begin{tikzpicture}
      \node[anchor=north west] (input) {Case 50 CT findings:};
      \node[below=0 of input.south west, draw, anchor=north west, text width = .72 \textwidth, xshift=1em] (input text) {\vbox{\ttfamily A 4.2 cm tumor is observed in the right upper lobe, surrounded by lung. No lymph node enlargement. No distant metastasis.}};
      \node[right=1em of input text, align=left]{Correct answer:\\T2b N0 M0};
      \path (0, 0 |- input text.south west) + (0, -1ex) coordinate (input bottom);
      \node[below=1ex of input bottom, anchor=north west] (ans) {Output by NotebookLM with REK:};
      \node[below=0 of ans.south west, draw, anchor=north west, xshift=1em, text width=.75 \textwidth] (nlm out) {\vbox{
        \ttfamily
        \begin{itemize}[leftmargin=*]
          \item \textbf{T2a:} The tumor has a solid component diameter greater than 3 cm and less than or equal to 4 cm, specifically 4.2 cm. This falls under the \textbf{T2} classification \textcircled{\scriptsize 1}. As the tumor size is 3 cm or less, it is classified as \textbf{T2a} \textcircled{\scriptsize 2}.
          \item \textbf{N0:} The case notes specify "no lymph node enlargement." This fits the criteria for \textbf{N0}, which signifies no regional lymph node metastasis \textcircled{\scriptsize 3}.
          \item \textbf{M0:} The case explicitly states "no distant metastasis." Therefore, it is classified as \textbf{M0}, indicating the absence of distant metastasis \textcircled{\scriptsize 4}.
        \end{itemize}
      }};
      \node[right=1em of nlm out] {Incorrect};
    \end{tikzpicture}
    \caption{An experimental result where NotebookLM made an incorrect numerical comparison. The sources \textcircled{\scriptsize 1} to \textcircled{\scriptsize 4} referenced in the answer are available in Online Resource 3\protect\footnotemark[2] (which happened to be the same as the case in \cref{repr}). \textit{REK}=reliable external knowledge.}
    \label{error}
  \end{figure}

  \section{Results}
  In the experiment using 100 fictional lung cancer cases, NotebookLM with REK diagnosed TNM classifications with a high accuracy of \qty{86}{\percent}, whereas GPT-4o with REK showed a lower diagnostic accuracy of \qty{39}{\percent}, and without REK, it dropped further to \qty{25}{\percent} (\cref{fig2}). \footnotetext[2]{Online Resources 1 to 3 can be found in the ancillary files uploaded with this paper on arXiv.}In contrast to GPT-4o with and without REK, NotebookLM explicitly presented the reference locations it searched within the REK as the basis for its diagnoses (see \cref{repr} for an example), and its search accuracy, representing the appropriateness of these locations, was quite high at \qty{95}{\percent} (\cref{fig2}). A similar trend was observed in the diagnostic accuracies for each of the T, N, and M factors, and in particular, NotebookLM's diagnostic accuracy for the T factor was notably higher than that of GPT-4o with and without REK (\cref{fig3}). Even so, with the high accuracy of NotebookLM, there were a few cases where, similar to GPT-4o, NotebookLM made errors in numerical comparisons (see \cref{error} for an example).

  \section{Discussion}
  There have been several previous studies on lung cancer stage classification using traditional LLMs, including GPT-4o, but their diagnostic accuracy is not considered high enough for clinical application \cite{Nakamura,Matsuo,Lee}. Similarly, in our experiment, the accuracy of lung cancer staging by GPT-4o was found to be insufficient (\cref{fig2,fig3}). As the guidelines for lung cancer TNM classification in various countries are frequently updated, the difficulty in determining which standards to reference from the vast amount of online information may be one reason for the low diagnostic accuracy of traditional LLMs.

  To improve the diagnostic accuracy of LLMs, we provided the latest lung cancer staging guideline in Japan as REK to the LLMs. As seen in \cref{fig2,fig3}, providing GPT-4o with the REK in its prompt resulted in slightly better performance compared to when the REK was not provided, but the diagnostic accuracy was still not sufficient. On the other hand, NotebookLM with REK demonstrated remarkably high diagnostic accuracy. This is likely because NotebookLM is designed to generate responses exclusively based on the provided REK through RAG, avoiding irrelevant information. 

  In our experiment, unlike GPT-4o, NotebookLM was able to clearly indicate the reference locations it found within the REK (see \cref{repr} for an example), and its search accuracy was notably high (\cref{fig2}). Although the search accuracy was slightly short of \qty{100}{\percent}, at \qty{95}{\percent}, the explicit indication of reference locations allows radiologists to easily verify the correctness of NotebookLM's responses as needed, making it a promising tool for providing reliable assistance in clinical diagnosis.

  Among the cases where the reference locations identified within the REK by NotebookLM were appropriate, there were a few instances of diagnostic errors, primarily caused by numerical comparison mistakes (see \cref{error} for an example). Such numerical reasoning errors were observed in GPT-4o as well and are recognized as a common issue with LLMs in general \cite{Li}. This is not an issue that can be resolved by RAG, and we hope that future advancements in LLMs will provide a solution.

  This study has several limitations. First, unlike in actual clinical settings, we validated the diagnostic accuracy of LLMs using fictional lung cancer CT findings and Japan's lung cancer staging guideline in English (instead of Japanese). To rigorously assess their usefulness in real clinical practice, actual clinical data should be used. Additionally, while this study demonstrated the usefulness of NotebookLM as a diagnostic support tool, it is not appropriate to generalize the results to all RAG-LLMs, and further validation using other RAG-LLMs is necessary for the generalization. Moreover, NotebookLM is currently a free online system in trial release by Google, available for easy use by anyone, but how it will evolve in the future remains uncertain. Furthermore, using online LLMs like GPT-4o and NotebookLM in clinical practice poses substantial challenges regarding hospital information security and copyright issues surrounding REK. To address this concern, it is likely that LLM systems for clinical use will need to be implemented offline or on-premises. Despite these limitations, we hope that this study will serve as a first step toward applying RAG-LLMs in image diagnosis.

  \section{Conclusion}
  NotebookLM, a recently released RAG-LLM from Google, demonstrated superior accuracy in a lung cancer staging experiment compared to GPT-4o, a widely recognized LLM. Additionally, NotebookLM explicitly provided accurate reference points from the given REK, allowing radiologists to efficiently assess the reliability of its responses and identify possible hallucinations. Overall, this study highlights the potential of NotebookLM in assisting radiologists with image diagnosis.

  \section{Declarations}
  \subsection*{Funding}
  This study was partially supported by JSPS KAKENHI Grant Number JP21K15762.
  \subsection*{Competing interests}
  There is no conflict of interest with regard to this manuscript.
  \subsection*{Ethics approval}
  Since the study used only fictional patient data, ethical approval was not required.
  \subsection*{Informed consent}
  Not applicable.
  \subsection*{Data availability statement}
  The data that support the findings of this study are available from the corresponding author, upon reasonable request.

  \bibliographystyle{unsrt}
  \bibliography{manuscript}
\end{document}